\pdfoutput=1

\documentclass[11pt]{article}

\usepackage{acl}

\usepackage{times}
\usepackage{latexsym}
\usepackage[T1]{fontenc}
\usepackage[utf8]{inputenc}
\usepackage{microtype}
\usepackage{inconsolata}
\usepackage{graphicx}
\usepackage{caption}
\usepackage{enumitem}
\usepackage{xurl}
\usepackage{xcolor}

\usepackage{listings}
\definecolor{bg}{rgb}{0.98,0.98,0.98}
\usepackage[multiple]{footmisc}
\title{Training a Huggingface Model on AWS Sagemaker (Without Tears)}

\author{Liling Tan}

\begin{document}
\maketitle
\begin{abstract}
Developments of Large language models (LLMs) have been mainly led by resource-rich research groups and industry partners. Without the luxury of on-premise computing resources to train LLMs of growing complexity, researchers are adopting cloud machine learning services like AWS Sagemaker to train Huggingface models. While a competent researcher can train a model well on a local machine, AWS Sagemaker's learning curve might be more challenging for non-seasoned cloud users. As a result, researchers end up scouring the web for information to fix issues and knowledge gaps in existing documentation to use Huggingface models in AWS Sagemaker. In this demo paper, to bring broader democratization of cloud adoption, we put the information to a single point such that research can start from zero to train their first Huggingface model. 
\end{abstract}

\section{Introduction}
State-of-the-art large language models (LLMs) have grown in size and training difficulty \citep{scao2022bloom,le-scao-etal-2022-language,smith2022using,alpaca}. While 90\% of the global IT spending is still on-premises and yet to migrate to the cloud \citep{andyjassy2023}, many research teams without heavy funding support do not have the luxury of on-premise computing cluster to train an LLM from scratch nor fine-tune a competitive state-of-the-art model with physical GPU machines. 

Unlike bare-metal cloud instances, products like Amazon Web Services (AWS) Sagemaker, Google Colaboratory, and Microsoft Azure ML Studio provide fuss-free access to Jupyter notebooks on managed virtual servers. These ``Jupyter as a Service" products abstract the users away from infrastructure setup and let researchers focus on data and model improvements with minimal Infra-/DevOps (Infrastructure or Development Operations) knowledge \citep{10.1145/3093338.3104159,perkel2018jupyter}.

Machine Learning Operationalization (MLOps) is the process of bringing machine learning models into production \citep{app11198861,huyen2022designing,10081336}. An end-to-end MLOps framework on the cloud would start from research and development tools support like ``Jupyter as a service" and the ability to measure, monitor and maintain the models created from researchers' notebooks in an automated manner. 

\begin{figure}[htp!]
 \centering
 \includegraphics[width=.8\columnwidth]{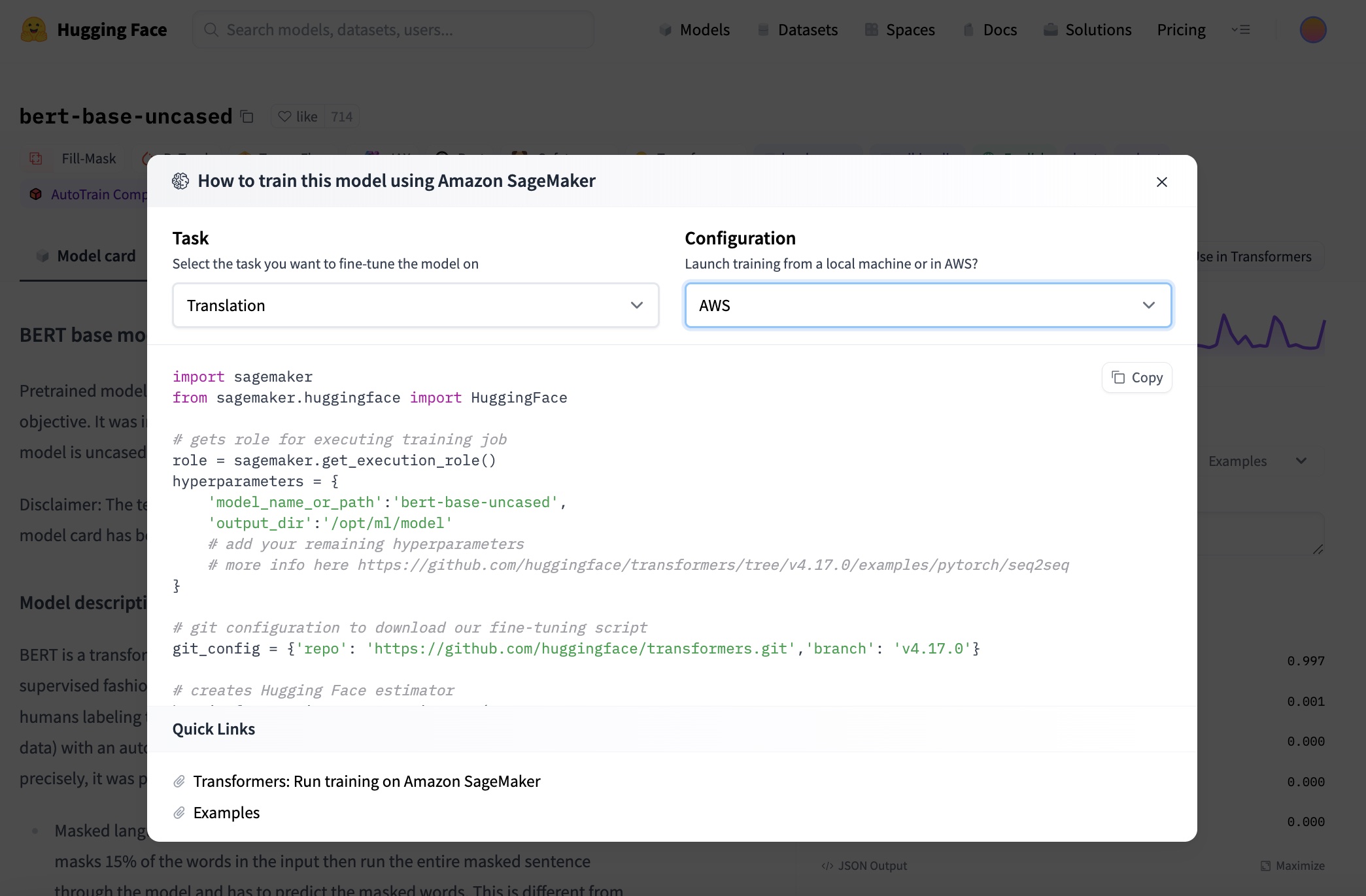}
 \caption{Sample code to train a Huggingface model using AWS Sagemaker}
 \label{hfaws}
\end{figure}

While there are a lot of books, documentation, blog posts, and a canonical online course\footnote{\url{https://huggingface.co/docs/sagemaker}} to educate users on HuggingFace with AWS Sagemaker, there's a lack of a comprehensive guide to training a HuggingFace model for new cloud users. Figure \ref{hfaws} shows the options on the Huggingface model card pages to create sample code to train a model using AWS Sagemaker. The sample code is helpful as a start, but researchers soon find that the access to features in Sagemaker is different from how one would code the same model on a local machine. 

This demo paper closes the knowledge gap between training a model on a local machine's Jupyter Notebook and training on AWS Sagemaker. We will present a demo to train a HuggingFace model using Huggingface on AWS Sagemaker. 

\section{AWS Sagemaker}

AWS Sagemaker is a managed cloud specifically used for machine learning. It comes with a suite of tools that ranges from data processing, Jupyter Notebook hosting, model provenance management, model training, hyperparameter tuning, and model inference hosting. In this demo, we will focus on model training and hyperparameter tuning and briefly touch on model deployment for the models we've trained in AWS Sagemaker. 

\subsection{Amazon SageMaker > Domains > Domain: mtsdemo > Jupyter Notebook}

To begin, a researcher would navigate to the AWS Sagemaker tool on the search bar in the AWS console. However, unlike tools like Google Colab and Kaggle Notebook, the user is not brought directly into a Jupyter Notebook but is required to \texttt{\textbf{Setup a Sagemaker Domain}} to configure permissions and security controls. 

\begin{figure}[htp!]
 \centering
 \includegraphics[width=.8\columnwidth]{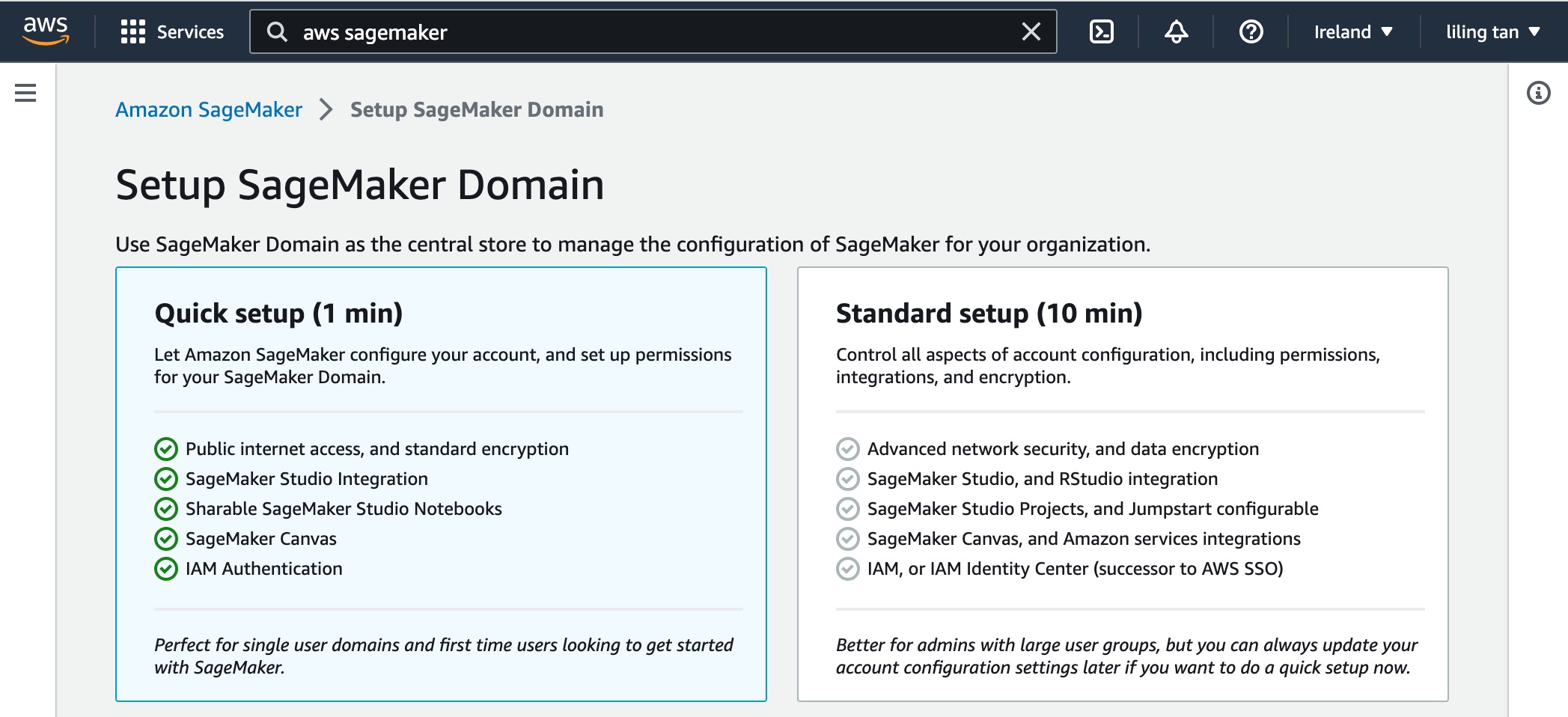}
 \captionof*{figure}{}
 \label{setupdomain}
\end{figure}
\vspace{-10mm}

For simplicity, a new user should use the quick setup; it takes 5-10 mins for the domain to be ready after the setup, after which it can be accessed from the left panel of the console through the \texttt{Domains} menu.

\begin{figure}[htp!]
 \centering
 \includegraphics[width=.7\columnwidth]{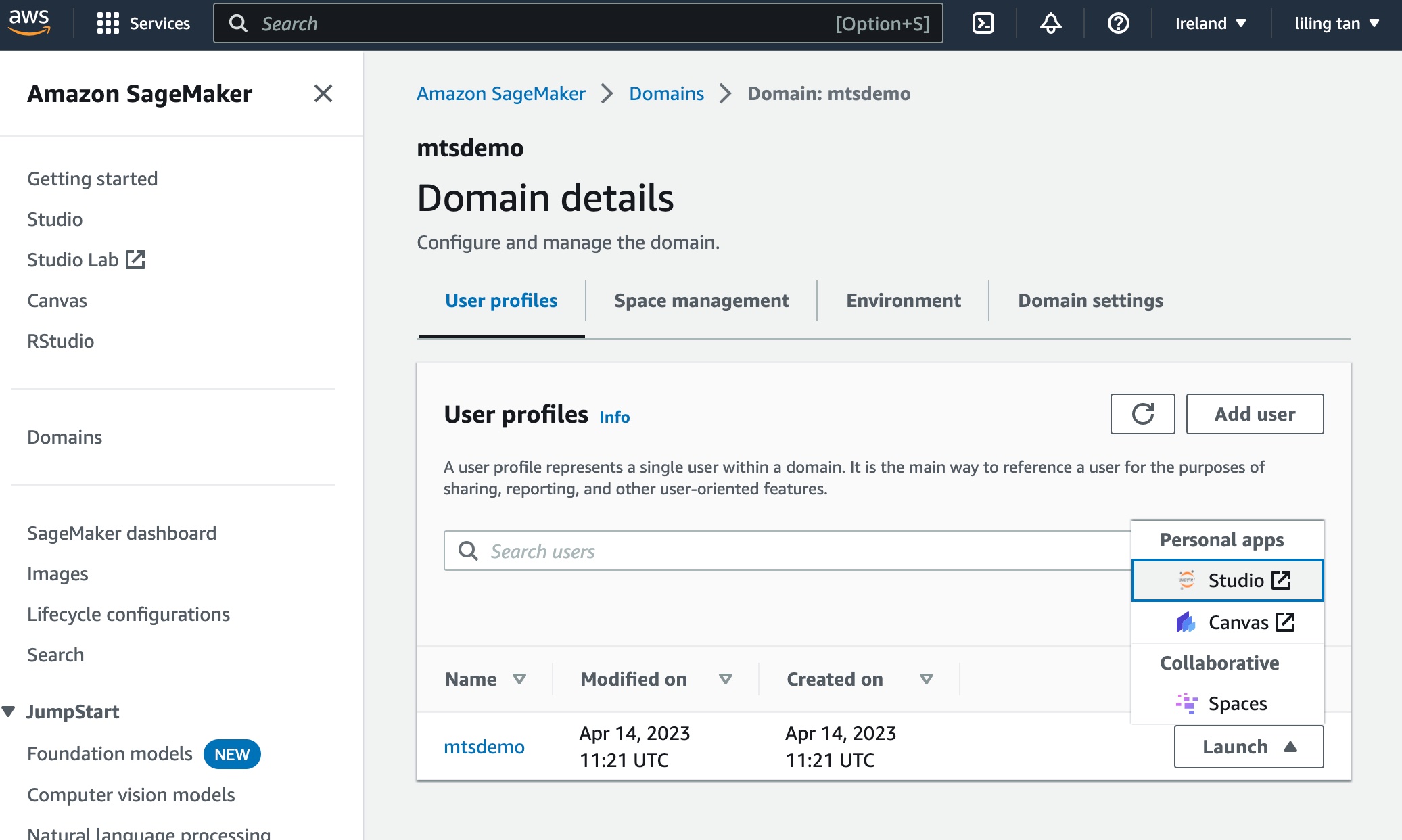}
 \captionof*{figure}{}
 \label{startstudio}
\end{figure}
\vspace{-10mm}

After clicking on the domain name, e.g., `mtsdomain`, there will be an option to open a \texttt{\textbf{Studio}} page. And that will bring the user to Sagemaker Studio's launcher page similar to \texttt{\textbf{Jupyter Lab}} on a local machine.

Next, on the top bar of the page, the user will need to create a new Jupyter notebook file using the \texttt{\textbf{File > New > Notebook}} and that will pop-up a \texttt{\textbf{Set up notebook environment}} window and user should configure the (i) image, (iii) instance type and (ii) kernel settings that fit the needs accordingly.

\begin{figure}[htp!]
 \centering
 \includegraphics[width=.8\columnwidth]{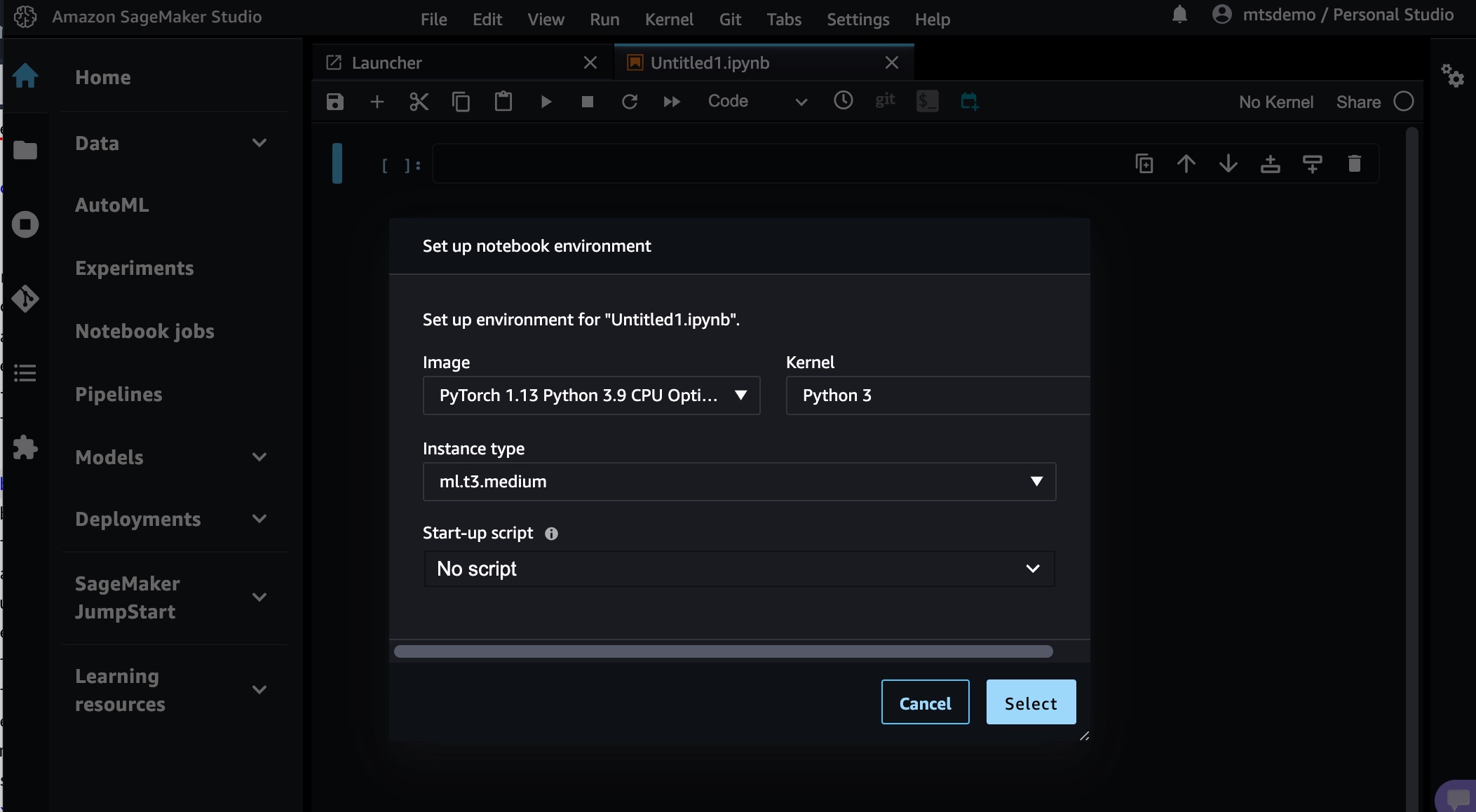}
 \label{setupnotebook}
\end{figure}

\begin{itemize}
\itemsep0em 
    \item \textbf{Image:} This is the Docker image from which you want to load your notebook.\footnote{If unsure, choose the latest CPU-optimized PyTorch and Python version compatible with the Huggingface transformer latest major/stable version}
    \item \textbf{Instance:} This is the instance type that you would like your Jupyter notebook to be hosted on. Note that this is not the instance type that you will train your model on in this demo, so a CPU instance works well.\footnote{If unsure, choose \texttt{ml.t3.medium}, the comprehensive list of instance can be found on \url{https://aws.amazon.com/ec2/instance-types/}}
    \item \textbf{Kernel:} It is always Python since we work with Huggingface libraries.
\end{itemize}

Finally, after setting up the notebook environment, users will see the familiar Jupyter notebook page as one would experience on a local machine, Google Colab, or Kaggle notebook.

\begin{figure}[htp!]
 \centering
 \includegraphics[width=.8\columnwidth]{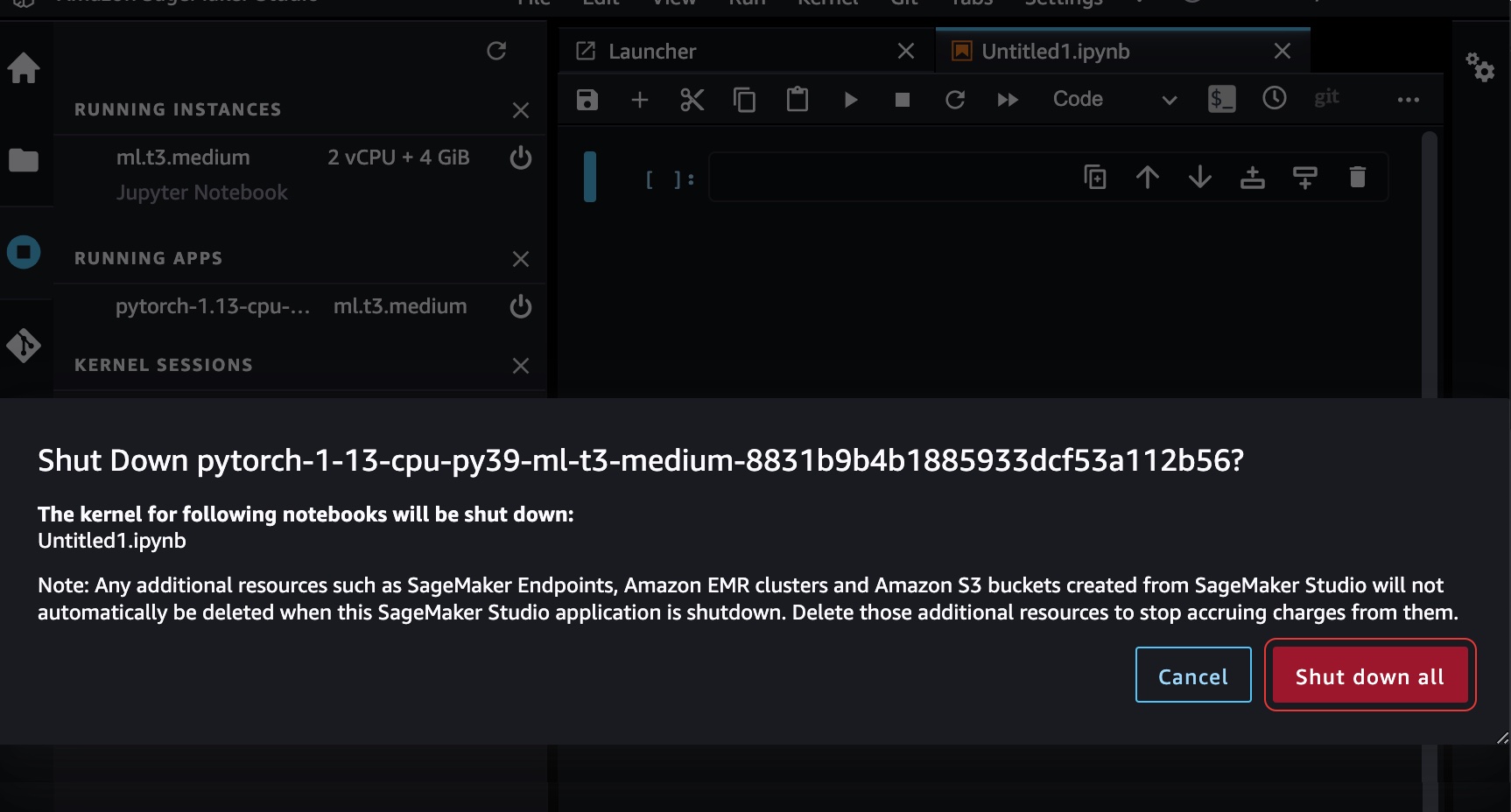}
 \caption{Shutting down virtual machine instances hosting the Jupyter notebook(s)}
 \label{shutdown_fig}
\end{figure}

After the Jupyter notebook runs, the charges to host the \texttt{ml.t3.medium} instance will be initiated, even if the notebook is not running anything. Stopping the notebook functions or ending the Jupyter runtime does not stop the charges since the virtual machine\footnote{For experienced cloud users, AWS calls them ``\textit{Amazon Elastic Compute Cloud Instances}", Microsoft Azure calls them ``\textit{Azure Virtual Machines}", Google Cloud Platform (GCP) calls them ``\textit{Compute Engine: Virtual Machines}". We will refer to virtual hardware that AWS Sagemaker uses as ``\textit{instances}" and ``\textit{virtual machine}" interchangeably} is still running and hosting notebook.

To stop the charges, frugal users are advised to open on the \texttt{instance, kernel, apps, and terminal session control} panel on the left bar, then click on the power-off button in the \textbf{\texttt{RUNNING INSTANCES}} and a pop-up will warn the user that shutting down the instances will stop all notebooks running on that instance and all unsaved data will be lost.\footnote{Note that there can be multiple notebook kernels running on the same instance, esp. when you select the same instance type when starting a new notebook. As a good practice, turn off the \texttt{KERNEL SESSIONS} you do not need before shutting down the \texttt{RUNNING INSTANCES} or \texttt{RUNNING APPS}}

The instructions in this section might seem trivial to seasoned AWS and Sagemaker users, but it can be daunting to new users and students learning to use cloud resources. While the instructions and user interface (UI) will change over time, the concepts on Jupyter services in MLOps remain; there will be some (i) hardware (\textit{virtual machine}) hosting the notebook, i.e., the compute resources needed to run a Jupyter notebook and (ii) software (\textit{Docker image}), i.e., the operating system and libraries that we will use to run the Jupyter notebook. 

In the case of AWS Sagemaker, the controls of the (i) and (ii) falls on the user's configurations, but with tools like Google Colab and Kaggle, they assume some defaults that most users will use and start with fixed (i) and (ii) configurations. 

\subsection{Amazon SageMaker > Training Jobs > Model Training} \label{quota}

After setting the Sagemaker Studio and having a Jupyter notebook running in the studio, users would expect the Huggingface example from Figure \ref{hfaws} to work out of the box using the HuggingFace estimator. But, there is one more hardware setup before model training. 

Unlike training a model on a local machine, where the machine running the Jupyter notebook is usually the same as the machine used to train the model, the instances that Sagemaker uses to run the Jupyter notebooks are generally not the machines that are used to train a model. One can think of the instance hosting the Jupyter Notebook as a controller machine that will turn on another machine or cluster of machines to start training the model, so the hosting for the notebook and the model training has different settings. 

The number of instances permitted to host Jupyter Notebook and run model training are limited as per the default quota set by AWS. For first-time Sagemaker users, you would first need to request a quota increase to use a GPU instance for the model training.

\begin{figure}[htp!]
 \centering
 \includegraphics[width=.8\columnwidth]{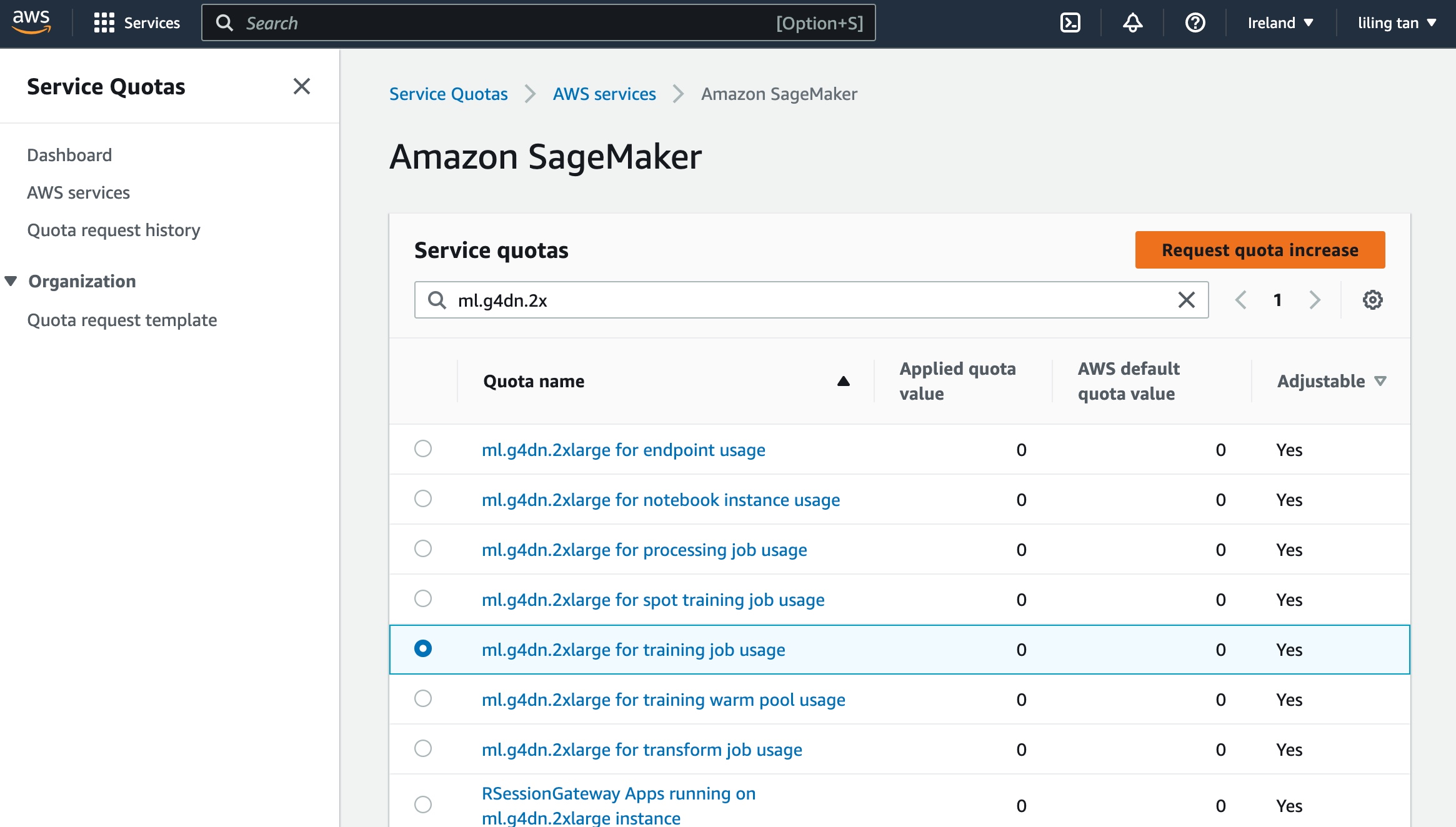}
 \label{quotarequest}
\end{figure}

User has to navigate to the \texttt{\textbf{Service Quotas > AWS services > Amazon SageMaker}} page, then type in \textit{Find quotas} search bar the instance type you want to train on, e.g., \texttt{ml.g4dn.2xlarge} and select the \textbf{\texttt{ml.g4dn.2xlarge} for training job usage} option to enable the \textbf{Request Quota Increase} button, follow through the quota increase request pop-up and increase the quota as desired.\footnote{If unsure, set the quota to 4; that means you can have a maximum of 4 concurrent \texttt{ml.g4dn.2xlarge} instance when training, users can use multiple instances to train a model or use multiple instances for a parallel hyperparameter grid search.}\textsuperscript{,}\footnote{Users can refer to \textbf{Accelerated Computing} section on \url{https://aws.amazon.com/ec2/instance-types/} to find the appropriate instance type; for references, the \texttt{P3} and \texttt{P4} instances come Nvidia V100 and A100 GPUs and \texttt{G4dn} instance comes with T4 GPUs} 

It will take a couple of hours or days to see the updated quota before users can train a model with the Sagemaker HuggingFace estimator.\footnote{Tip: From experience, the fastest approval for GPU instances comes from requesting only 1 \texttt{ml.g4dn.xlarge} instance for training job usage.} The basic quota request on the platform through the automated \textbf{Service Request} page does not incur any charges, but to expedite or escalate the quota request with a tech support personnel, users can subscribe to various tiers of AWS Support Plans at additional cost. Users can track the status of the quota request from the \textbf{Service Quotas > Quota request history} on the left panel of the page.

\subsubsection{Your First Huggingface Script on AWS Sagemaker}

After getting the quota request approved, users can use the sample code (with slight modification) in Figure \ref{hfaws} to train their first Huggingface model on AWS Sagemaker. However, the verbatim sample code will not work out of the box since some task and model-specific hyperparameters need to be initialized.

Figure \ref{working-sample-fig} (in Appendix) shows a working script adapted from the Huggingface sample code to launch a Python script that uses the \texttt{transformers} library with the \texttt{sagemaker.huggingface.HuggingFace} model estimator. 

\begin{figure}[htp!]
 \centering
 \includegraphics[width=.975\columnwidth]{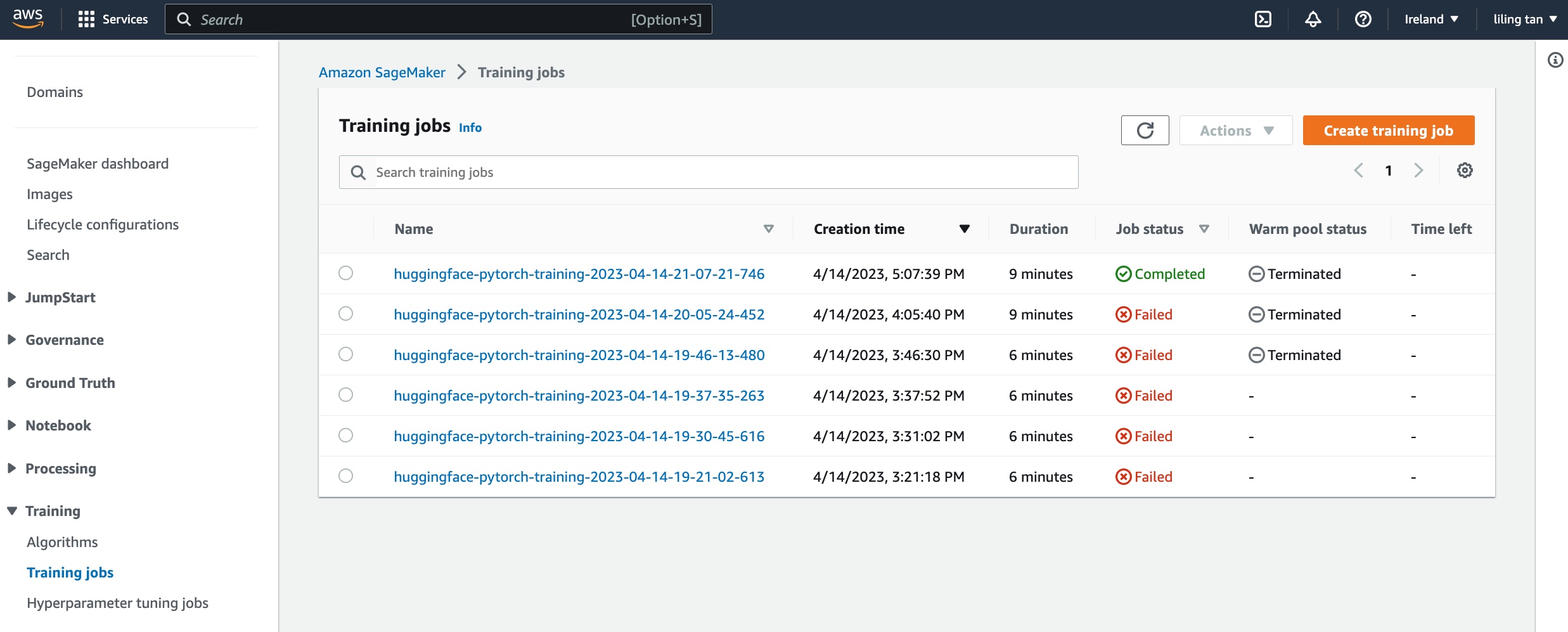}
 \caption{Monitoring Model Trainings on the AWS Sagemaker Console}
 \label{trackmodel}
\end{figure}

While the script is running, users can navigate to the \texttt{\textbf{Amazon SageMaker > Training > Training jobs}} page on the left panel and track the progress of the model training. For example, Figure \ref{trackmodel} shows the dashboard of the successful/failed runs of the scripts that were launched using the \texttt{sagemaker.huggingface.HuggingFace} estimators as presented in Figure \ref{working-sample-fig}.

Up till this point, pointers and procedures presented in the above sections were unavailable in a single point of documentation; bits and pieces lurk around open tutorials, demonstrations and online courses. After a user manages to get a script shown in Figure \ref{working-sample-fig} working, there are many other open and accessible demonstrations to use the various Huggingface libraries with AWS Sagemaker. Notably, we refer readers to start with \url{https://huggingface.co/docs/sagemaker/train} for further reading on the generic use of Huggingface models with AWS Sagemaker. 

While documentation to fine-tune \texttt{transformers} models with preset datasets available on Huggingface \texttt{datasets} library are widely available, users are frequently asking `\textit{How do I train Huggingface model X with custom data Y with AWS Sagemaker?}' questions on the official Sagemaker section of the Huggingface community forum and StackOverflow's \texttt{huggingface-transformers} and \texttt{aws-sagemaker} tagged questions.\footnote{\url{https://stackoverflow.com/questions/tagged/huggingface-transformers\%20amazon-sagemaker}}\textsuperscript{,}\footnote{\url{https://discuss.huggingface.co/c/sagemaker/17}}

For the rest of this demo paper, we will explain the sample code from Figure \ref{working-sample-fig} and walk through the code line-by-line with the goal of empowering users to understand what they mean and how they should modify them to their specific NLP needs. As we explain the script, hopefully, it will answer Frequently Asked Questions (FAQs) on using Huggingface libraries with AWS Sagemaker. We will also highlight lesser-known AWS Sagemaker features to maximize developers' productivity by emulating a researchers' process to use a Jupyter Notebook on non-managed bare-metal virtual machines.

\section{HuggingFace Estimator}

We begin the script walkthrough starting from the \texttt{HuggingFace} estimator:

\begin{lstlisting}[language=Python]
git_config = {
  'repo': 'https://github.com/'
    'huggingface/transformers.git',
  'branch': 'v4.36.0'}
  
huggingface_estimator = HuggingFace(
  # Code location arguments.
  git_config=git_config,
  entry_point='run_translation.py',
  source_dir='./examples/'
             'pytorch/translation',
  # AWS instance arguments.
  role=role,
  instance_type='ml.g4dn.xlarge',
  instance_count=1,
  # Docker image arguments.
  transformers_version='4.36.0',
  pytorch_version='2.1.0',
  py_version='py310',
  # Model's hyperparameters.
  hyperparameters = hyperparameters,
)
\end{lstlisting}

Other than the model's hyperparameters that are dependent on how users coded the argument parser in their training script, there are three groups of Sagemaker-specific keyword arguments that the \texttt{HuggingFace} estimator expects (i) code location, (ii) AWS instance and (iii) Docker image arguments.

\subsection{Specifying Code Location for \texttt{HuggingFace} Estimator}

Looking at the \texttt{git\_config} variable initialized before the \texttt{HuggingFace} estimator, we see that it is a Python dictionary pointing to the public GitHub repository \url{https://github.com/huggingface/transformers.git} and directing the estimator to pull the code from the \texttt{v4.36.0} branch on the git tree. 

The \texttt{git\_config} together the \texttt{source\_dir} and \texttt{entry\_point} is asking the \texttt{HuggingFace} estimator to pull the \texttt{run\_translation.py} script.

\subsubsection*{Q: Pulling from GitHub is nice, but how do I use my own Python script?}

To train from a local script, specify the path to the local directory containing your \texttt{.py} file. For example, if you have a \texttt{script/my\_run\_translation.py} file in the same directory as Jupyter notebook that you are running the \texttt{HuggingFace} estimator, e.g., \texttt{train\_hf\_sagemaker.ipynb},

\begin{lstlisting}[language=Python]
  /script
     my_run_translation.py
  train_hf_sagemaker.ipynb
\end{lstlisting}

Then users can remove the \texttt{git\_config} variable and use the \texttt{source\_dir} and \texttt{entry\_point} arguments as follows:

\begin{lstlisting}[language=Python]
huggingface_estimator = HuggingFace(
  # Code location arguments.
  entry_point='my_run_translation.py',
  source_dir='./script',
  # AWS instance arguments.
  role=role,
  instance_type='ml.g4dn.xlarge',
  instance_count=1,
  # Docker image arguments.
  transformers_version='4.36.0',
  pytorch_version='2.1.0',
  py_version='py310',
  # Model's hyperparameters.
  hyperparameters = hyperparameters,
)
\end{lstlisting}

There are other modes of utilizing the \texttt{git\_config} argument of the estimator object; we refer readers to \citet{engdahl2008} comprehensive tutorial to integrate other GitHub repository or AWS CodeCommit.

\subsubsection*{Q: What if my custom training script uses libraries that are not supported by Huggingface \texttt{transformers} natively?}

The simplest solution is a somewhat hidden AWS Sagemaker feature where you can enforce a \texttt{\textbf{SAGEMAKER\_ENVIRONMENT}} variable that contains the \texttt{requirements.txt} file that one would usually use with Python \texttt{pip} in the same directory as the \texttt{source\_dir}. For example, with a directory structure as follows,

\begin{lstlisting}[language=Python]
  /script
     my_run_translation.py
     requirements.txt
  train_hf_sagemaker.ipynb
\end{lstlisting}

Users can specify the \texttt{env} argument, and Sagemaker will install the Python packages listed in the \texttt{requirements.txt} after starting up the virtual machine. 

\begin{lstlisting}[language=Python]
huggingface_estimator = HuggingFace(
  # Code location arguments.
  entry_point='my_run_translation.py',
  source_dir='./script',
  env={'SAGEMAKER_ENVIRONMENT': 
         'requirements.txt'},
  # AWS instance arguments.
  role=role,
  instance_type='ml.g4dn.xlarge',
  instance_count=1,
  # Docker image arguments.
  transformers_version='4.36.0',
  pytorch_version='2.1.0',
  py_version='py310',
  # Model's hyperparameters.
  hyperparameters = hyperparameters,
)
\end{lstlisting}

The snippets above with the fully functional code on Figure \ref{working-sample-fig} (in Appendix) is sufficient to run an actual model training. There are more tips and tricks on \textbf{Controlling AWS Instance for HuggingFace Estimator} and \textbf{Hyperparameter tuning jobs with Sagemaker} in Appendix B and C. 

\section{Come for the Code, Stay for the Community}

Open source and open knowledge documentation are fueled by the community. NLP researchers and ML practitioners are pragmatic, but for any open source to be successful, people come for the code and stay in the community \citep{naomiclsc,drupalleander}. 
The content presented in this demo is only possible with the bits and pieces of knowledge that many developers and researchers have recorded from the various forums. 

As a final note to the readers, we want to encourage everyone to openly discuss and post and/or answer questions on the canonical forum for the Huggingface x AWS Sagemaker topic, \url{https://discuss.huggingface.co/c/sagemaker/17}. We hope to see the knowledge shared here improves with similar open documentation efforts.

\section{Related Work}
Unlike a typical demonstration paper, we are not presenting an NLP tool with a results table we can easily compare with other related work. Thus, we highlight several works introducing NLP tooling paradigms as further references to related NLP and MLOps literature.

\citet{kim-etal-2021-changes} briefly described a no-code or low-code NLP experimentation setup to tune pre-trained models through an API without explicitly coding up the tuning procedures. While there seems to be little scientific contribution for researchers using AutoML, we concur with \citet{kim-etal-2021-changes} on how low-/no-code setups are useful first-phase proof-of-concept tools before investing time and resources into coding up a script and running it on a Jupyter-as-a-service on the cloud.

Another line of work describes the data-centric MLOps contributions, notably, \cite{lhoest-etal-2021-datasets} introduced the HuggingFace datasets hub that allows researchers to contribute, share and read datasets; their focus on API standardization, ease-of-use and community-drive data statement documentation \citep{bender-friedman-2018-data} has springboard a new generation of open datasets sharing and allows open-ended discussion on biased or harmful data artifacts \cite{ramponi-tonelli-2022-features}.

En vogue, Jupyter is the tool of choice to write NLP code and interact with the models \citep{perkel2018jupyter}. The closest comparable resources to our paper on model training models beyond Jupyter notebook are \citet{beyondjupyter} introduction to environment setup and the HuggingFace Transformers with Sagemaker tutorial. However, they did not cover the various issues and resolutions that users will face when new to AWS Sagemaker.

\section{Conclusion}

We have demonstrated various tips and tricks to bridge the knowledge gaps using the Huggingface estimators on the AWS Sagemaker cloud service. Bringing the information from different blog posts, tutorials, and QnAs sites into this demo paper, we hope readers have a comprehensive guide to kick-start their Huggingface x AWS Sagemaker usage. Although this demo paper is not a traditional format of presenting a tool concerning results and tools, we introduced the engineering concepts in MLOps and practical steps to apply them in training NLP models. 


\bibliography{custom}

@article{scao2022bloom,
  title={Bloom: A 176b-parameter open-access multilingual language model},
  author={Scao, Teven Le and Fan, Angela and Akiki, Christopher and Pavlick, Ellie and Ili{\'c}, Suzana and Hesslow, Daniel and Castagn{\'e}, Roman and Luccioni, Alexandra Sasha and Yvon, Fran{\c{c}}ois and Gall{\'e}, Matthias and others},
  journal={arXiv preprint arXiv:2211.05100},
  year={2022}
}

@inproceedings{le-scao-etal-2022-language,
    title = "What Language Model to Train if You Have One Million {GPU} Hours?",
    author = "Le Scao, Teven  and
      Wang, Thomas  and
      Hesslow, Daniel  and
      Bekman, Stas  and
      Bari, M Saiful  and
      Biderman, Stella  and
      Elsahar, Hady  and
      Muennighoff, Niklas  and
      Phang, Jason  and
      Press, Ofir  and
      Raffel, Colin  and
      Sanh, Victor  and
      Shen, Sheng  and
      Sutawika, Lintang  and
      Tae, Jaesung  and
      Yong, Zheng Xin  and
      Launay, Julien  and
      Beltagy, Iz",
    booktitle = "Findings of the Association for Computational Linguistics: EMNLP 2022",
    month = dec,
    year = "2022",
    address = "Abu Dhabi, United Arab Emirates",
    publisher = "Association for Computational Linguistics",
    url = "https://aclanthology.org/2022.findings-emnlp.54",
    pages = "765--782",
}

@misc{smith2022using,
      title={Using DeepSpeed and Megatron to Train Megatron-Turing NLG 530B, A Large-Scale Generative Language Model}, 
      author={Shaden Smith and Mostofa Patwary and Brandon Norick and Patrick LeGresley and Samyam Rajbhandari and Jared Casper and Zhun Liu and Shrimai Prabhumoye and George Zerveas and Vijay Korthikanti and Elton Zhang and Rewon Child and Reza Yazdani Aminabadi and Julie Bernauer and Xia Song and Mohammad Shoeybi and Yuxiong He and Michael Houston and Saurabh Tiwary and Bryan Catanzaro},
      year={2022},
      eprint={2201.11990},
      archivePrefix={arXiv},
      primaryClass={cs.CL}
}

@misc{alpaca,
  author = {Rohan Taori and Ishaan Gulrajani and Tianyi Zhang and Yann Dubois and Xuechen Li and Carlos Guestrin and Percy Liang and Tatsunori B. Hashimoto },
  title = {Stanford Alpaca: An Instruction-following LLaMA model},
  year = {2023},
  publisher = {GitHub},
  journal = {GitHub repository},
  howpublished = {\url{https://github.com/tatsu-lab/stanford_alpaca}},
}

@misc{andyjassy2023, title={CEO Andy Jassy's 2022 letter to shareholders}, url={https://www.aboutamazon.com/news/company-news/amazon-ceo-andy-jassy-2022-letter-to-shareholders}, Howpublished={About Amazon}, publisher={ About Amazon}, author={Andy Jassy}, year={2023}, month={Apr}}

@book{huyen2022designing,
  title={Designing Machine Learning Systems},
  author={Huyen, Chip},
  year={2022},
  publisher={O'Reilly Media, Inc.}
}

@Article{app11198861,
AUTHOR = {Ruf, Philipp and Madan, Manav and Reich, Christoph and Ould-Abdeslam, Djaffar},
TITLE = {Demystifying MLOps and Presenting a Recipe for the Selection of Open-Source Tools},
JOURNAL = {Applied Sciences},
VOLUME = {11},
YEAR = {2021},
NUMBER = {19},
ARTICLE-NUMBER = {8861},
URL = {https://www.mdpi.com/2076-3417/11/19/8861},
ISSN = {2076-3417},
DOI = {10.3390/app11198861}
}

@article{perkel2018jupyter,
  title={Why Jupyter is data scientists' computational notebook of choice},
  author={Perkel, Jeffrey M},
  journal={Nature},
  volume={563},
  number={7732},
  pages={145--147},
  year={2018},
  publisher={Nature Publishing Group}
}

@ARTICLE{10081336,
  author={Kreuzberger, Dominik and Kühl, Niklas and Hirschl, Sebastian},
  journal={IEEE Access}, 
  title={Machine Learning Operations (MLOps): Overview, Definition, and Architecture}, 
  year={2023},
  volume={11},
  number={},
  pages={31866-31879},
  doi={10.1109/ACCESS.2023.3262138}}

@inproceedings{10.1145/3093338.3104159,
author = {Milligan, Michael},
title = {Interactive HPC Gateways with Jupyter and Jupyterhub},
year = {2017},
isbn = {9781450352727},
publisher = {Association for Computing Machinery},
address = {New York, NY, USA},
url = {https://doi.org/10.1145/3093338.3104159},
doi = {10.1145/3093338.3104159},
booktitle = {Proceedings of the Practice and Experience in Advanced Research Computing 2017 on Sustainability, Success and Impact},
articleno = {63},
numpages = {4},
location = {New Orleans, LA, USA},
series = {PEARC17}
}

@misc{engdahl2008, title={Git integration now available for the Amazon SageMaker Python SDK}, url={https://aws.amazon.com/blogs/machine-learning/git-integration-now-available-for-amazon-sagemaker-python-sdk/}, Howpublished={Amazon}, publisher={Amazon}, author={Yue Tu and Chuyang Deng}, year={2008}}

@misc{hotz2022, title={Unlock the latest transformer models with Amazon Sagemaker}, url={https://towardsdatascience.com/unlock-the-latest-transformer-models-with-amazon-sagemaker-7fe65130d993}, Howpublished={Medium Blogpost}, publisher={Towards Data Science}, author={Hotz, Heiko}, year={2022}, month={Dec}}

@misc{cloudmap, title={Cloud Product Mapping (AWS vs Azure vs GCP)}, url={https://github.com/milanm/Cloud-Product-Mapping}, Howpublished={GitHub Repository}, publisher={GitHub}, author={Milan Milanovic }, year={2022}, month={Dec}}

@misc{sabin2023, title={Hackers are quickly learning how to Breach Cloud Systems}, url={https://www.axios.com/2023/03/07/hackers-cloud-breaches-cybersecurity}, Howpublished={Axios }, author={Sabin, Sam}, year={2023}, month={Mar}}

@misc{thehackernews2023, title={Microsoft Azure services flaws could've exposed cloud resources to unauthorized access}, url={https://thehackernews.com/2023/01/microsoft-azure-services-flaws-couldve.html}, Howpublished={The Hacker News}, year={2023}, month={Jan}, author={Ravie Lakshmanan}}

@misc{drupalleander,
    title = {Drupal – come for the code, stay for the community},
    year = {2018},
    organization = {Foo Cafe},
    author = {Leander Lindahl and Jens Grip},
    url = {https://www.youtube.com/watch?v=FMUr4U8oZ_4},
    Howpublished = {Foo Cafe. Retrieved from \url{https://www.youtube.com/watch?v=FMUr4U8oZ_4}},
    urldate = {2023-04-15}
}

@misc{naomiclsc,
    title = {Come for the Language, Stay for the Community},
    year = {2016},
    organization = {Euro PyCon},
    author = {Naomi Ceder},
    url = {https://www.youtube.com/watch?v=cCCiA-IlVco},
    Howpublished = {Euro PyCon. Retrieved from \url{https://www.youtube.com/watch?v=cCCiA-IlVco}}}

@misc{heiligenstein2023, 
 title={Amazon Web Services (AWS) data breaches: Full timeline through 2023}, 
 url={https://firewalltimes.com/amazon-web-services-data-breach-timeline/}, journal={Firewall Times}, author={Heiligenstein, Michael X.}, year={2023}, month={Apr}, 
 Howpublished = {Firewall Times}

 }

@misc{pinto2011,
title={High-Performance Computing Needs Machine Learning... And Vice Versa}, 
url={https://www.slideshare.net/npinto/highperformance-computing-needs-machine-learning-and-vice-versa-nips-2011-big-learning},Howpublished ={NIPS 2011, Big Learning},
year=2011,
author={Nicolas Pinto}
}

@misc{beyondjupyter,
    title = {Beyond {Jupyter} {Notebooks}: {MLOps} Environment Setup \& First Deployment},
    author = {Greg Loughnane},
    year = {2022},
    howpublished={DeepLearning.AI}
}

@inproceedings{kim-etal-2021-changes,
    title = "What Changes Can Large-scale Language Models Bring? Intensive Study on {H}yper{CLOVA}: Billions-scale {K}orean Generative Pretrained Transformers",
    author = "Kim, Boseop  and
      Kim, HyoungSeok  and
      Lee, Sang-Woo  and
      Lee, Gichang  and
      Kwak, Donghyun  and
      Dong Hyeon, Jeon  and
      Park, Sunghyun  and
      Kim, Sungju  and
      Kim, Seonhoon  and
      Seo, Dongpil  and
      Lee, Heungsub  and
      Jeong, Minyoung  and
      Lee, Sungjae  and
      Kim, Minsub  and
      Ko, Suk Hyun  and
      Kim, Seokhun  and
      Park, Taeyong  and
      Kim, Jinuk  and
      Kang, Soyoung  and
      Ryu, Na-Hyeon  and
      Yoo, Kang Min  and
      Chang, Minsuk  and
      Suh, Soobin  and
      In, Sookyo  and
      Park, Jinseong  and
      Kim, Kyungduk  and
      Kim, Hiun  and
      Jeong, Jisu  and
      Yeo, Yong Goo  and
      Ham, Donghoon  and
      Park, Dongju  and
      Lee, Min Young  and
      Kang, Jaewook  and
      Kang, Inho  and
      Ha, Jung-Woo  and
      Park, Woomyoung  and
      Sung, Nako",
    booktitle = "Proceedings of the 2021 Conference on Empirical Methods in Natural Language Processing",
    month = nov,
    year = "2021",
    address = "Online and Punta Cana, Dominican Republic",
    publisher = "Association for Computational Linguistics",
    url = "https://aclanthology.org/2021.emnlp-main.274",
    doi = "10.18653/v1/2021.emnlp-main.274",
    pages = "3405--3424",
}

@inproceedings{lhoest-etal-2021-datasets,
    title = "Datasets: A Community Library for Natural Language Processing",
    author = "Lhoest, Quentin  and
      Villanova del Moral, Albert  and
      Jernite, Yacine  and
      Thakur, Abhishek  and
      von Platen, Patrick  and
      Patil, Suraj  and
      Chaumond, Julien  and
      Drame, Mariama  and
      Plu, Julien  and
      Tunstall, Lewis  and
      Davison, Joe  and
      {\v{S}}a{\v{s}}ko, Mario  and
      Chhablani, Gunjan  and
      Malik, Bhavitvya  and
      Brandeis, Simon  and
      Le Scao, Teven  and
      Sanh, Victor  and
      Xu, Canwen  and
      Patry, Nicolas  and
      McMillan-Major, Angelina  and
      Schmid, Philipp  and
      Gugger, Sylvain  and
      Delangue, Cl{\'e}ment  and
      Matussi{\`e}re, Th{\'e}o  and
      Debut, Lysandre  and
      Bekman, Stas  and
      Cistac, Pierric  and
      Goehringer, Thibault  and
      Mustar, Victor  and
      Lagunas, Fran{\c{c}}ois  and
      Rush, Alexander  and
      Wolf, Thomas",
    booktitle = "Proceedings of the 2021 Conference on Empirical Methods in Natural Language Processing: System Demonstrations",
    month = nov,
    year = "2021",
    address = "Online and Punta Cana, Dominican Republic",
    publisher = "Association for Computational Linguistics",
    url = "https://aclanthology.org/2021.emnlp-demo.21",
    doi = "10.18653/v1/2021.emnlp-demo.21",
    pages = "175--184",
}

@article{bender-friedman-2018-data,
    title = "Data Statements for Natural Language Processing: Toward Mitigating System Bias and Enabling Better Science",
    author = "Bender, Emily M.  and
      Friedman, Batya",
    journal = "Transactions of the Association for Computational Linguistics",
    volume = "6",
    year = "2018",
    address = "Cambridge, MA",
    publisher = "MIT Press",
    url = "https://aclanthology.org/Q18-1041",
    doi = "10.1162/tacl_a_00041",
    pages = "587--604",
}

@inproceedings{ramponi-tonelli-2022-features,
    title = "Features or Spurious Artifacts? Data-centric Baselines for Fair and Robust Hate Speech Detection",
    author = "Ramponi, Alan  and
      Tonelli, Sara",
    booktitle = "Proceedings of the 2022 Conference of the North American Chapter of the Association for Computational Linguistics: Human Language Technologies",
    month = jul,
    year = "2022",
    address = "Seattle, United States",
    publisher = "Association for Computational Linguistics",
    url = "https://aclanthology.org/2022.naacl-main.221",
    doi = "10.18653/v1/2022.naacl-main.221",
    pages = "3027--3040",
}

\section*{Appendix}
\appendix
\label{sec:appendix}

All the code snippets and Frequently Asked Questions listed on this paper will be hosted on \url{https://github.com/anonymous/hfxaws-no-tears}.

Like many NLP tools and deep learning library, the need to update the dependencies and code-base can be a never-ending pain. As the paper revision just before submitting the first draft to the conference, the minor versions of Huggingface transformers have changed at a monthly pace. We recommend the version controls with the solutions in Section 3.1 and 3.3. 

\section{Limitations and Risk}

\textbf{Limitations (Engineering): }This paper presented various knowledge gaps in training a Huggingface model on AWS Sagemaker. Although the content is hyper-limited to the Amazon Web Services cloud usage, the general MLOps tips presented in the form of AWS-related services are portable to other cloud providers (like Google Cloud Platform or Microsoft Azure) with their respective equivalent configurations. We refer readers to this \citet{cloudmap} that maps the cloud services across the providers.
\\ \\
\textbf{Limitations (Scientific): }Although this paper presented the procedures and information to launch a Huggingface model training job on AWS Sagemaker, we have largely skipped the description on how to code \texttt{entry\_point} script to build state-of-the-art NLP models. While this is in scope for the topics discussed in this paper, the code in the \texttt{entry\_point} script is the same as how a researcher would train a model on a local machine. We acknowledge that there are plenty of good resources with little knowledge gaps on that, esp. on the cannonical Huggingface documentation site. 
\\ \\
\noindent \textbf{Risk: } While there is no evident risk in the demonstrations presented in this paper. There are various security risks that cloud users should be concerned about. Using a self-contained on-premise environment, data stays within the organization's server. But with cloud access, there will always be certain risks to cloud breaches \citep{sabin2023,heiligenstein2023,thehackernews2023}; we advise users to consult their organizations' security administrators before sending private data onto the cloud.

\section{Controlling AWS Instance for HuggingFace Estimator}

As discussed in Section \ref{quota}, the quota issue will be the first blocker most users will face when setting the instance count. After increasing the quota and having the instance made available to Sagemaker for training jobs, users will soon find that only specific instance types are allowed to run the HuggingFace estimator. 

\subsubsection*{Q: Aren't the \texttt{instance\_type=} and \texttt{instance\_count} just boilerplate arguments?}

Yes, they are; if you have finished training your model in a reasonable time and everything works without GPU memory limitation issues. 

Suppose you encounter the cursed \texttt{CUDA Out of Memory} error, or the model is taking unreasonably long to converge. In that case, it is time to consider changing the AWS instances arguments to speed up or scale up your training jobs.

\subsubsection*{Q: How do I speed up or scale up my Huggingface model training with AWS Sagemaker?}

For beginner users, we suggest following the simple Data Parallelism as suggested on multi-GPUs instances, e.g., \texttt{ml.g4dn.2xlarge}.

\begin{lstlisting}[language=Python]
# configuration for running training 
# on smdistributed data parallel
distribution = {
  'smdistributed':{
    'dataparallel':{ 
      'enabled': True }
    }
}

huggingface_estimator = HuggingFace(
  # Code location arguments.
  entry_point='my_run_translation.py',
  source_dir='./script',
  # AWS instance arguments.
  role=role,
  instance_type='ml.g4dn.2xlarge',
  instance_count=1,
  # Docker image arguments.
  transformers_version='4.36.0',
  pytorch_version='2.1.0',
  py_version='py310',
  # Model's hyperparameters.
  hyperparameters = hyperparameters,
  distribution = distribution
)
\end{lstlisting}

And when the model exceeds the computes possible on a single instance, then user can consider other parallelism options native to AWS or third-party parallelism libraries integrated into Huggingface \texttt{transformers}.

The AWS documentation page keeps an updated list of FAQs about picking AWS instances configurations to speed up training and/or reduce cloud spending budgets. We refer users to these tutorials to optimize AWS instance controls using HuggingFace estimators.

\subsection{Selecting Docker Image for HuggingFace Estimator}

As per writing this demo paper, the sample Sagemaker training codes from the Huggingface model cards have been referring to the de facto Huggingface v4.36.0 and PyTorch v2.1.0. 

One of the issues that many users will encounter when setting the library versions to \texttt{transformers<=4.36.0} is that the model they want to tune/train on is created post v4.36.0, e.g., users were excited to start using the Open AI Whisper model, but it is only supported in later versions of the transformers library not supported by then latest Sagemaker Docker image available.

It takes a while for new Sagemaker users to figure out which versions are supported in the preset docker images from the Deep Learning Containers (DLC) before they find links that navigate to the canonical list of supported DLC. 

While the decision to make users explicitly state the transformers, pytorch, and python versions seems tedious instead of automatically fetching the latest image, it is a readability trade-off following the `\textit{explicit is better than implicit}' Pythonic \textit{satori}. 

\subsubsection*{Q: Can I Bring-Your-Own-Container (BYOC) to run the HuggingFace Estimator?}

Yes, you can. If the account you are accessing the Sagemaker notebook from has access to the docker image or if you are using some publicly available docker image, you can use the \texttt{image\_uri} instead of specifying the transformers, pytorch and Python versions. 

\begin{lstlisting}[language=Python]
ecr_uri=str(
"123456789012.dkr.ecr.eu-west-1."
"azmazonaws.com"
"/my-awsome-huggingface-docker:latest"
)

huggingface_estimator = HuggingFace(
  # Code location arguments.
  entry_point='my_run_translation.py',
  source_dir='./script',
  # AWS instance arguments.
  role=role,
  instance_type='ml.g4dn.2xlarge',
  instance_count=1,
  # Docker image arguments.
  image_uri=ecr_uri,
  # Model's hyperparameters.
  hyperparameters = hyperparameters
)
\end{lstlisting}

\citet{hotz2022} provided a detailed tutorial on modifying existing DLC images and modifying it before using them as a custom BYOC setup in AWS Sagemaker.

\section{Amazon SageMaker > Hyperparameter tuning jobs}

While many tutorials on using Sagemaker stop at instructing readers/students to train or deploy a Huggingface model, hyperparameter tuning is often left out, and users end up with manual tracking on different training runs.

Figure \ref{working-sample-fig} introduces a simple sample code to capitalize on the Automatic Model Tuning feature in Sagemaker and release users from performing ``\textit{graduate student descent }" \citep{pinto2011}. Alternatively, we encourage readers to explore the Syne Tune feature to perform more complex hyperparameter tuning tasks.

\newpage
\onecolumn
\begin{figure}[!ht]
\begin{lstlisting}[language=Python]
import sagemaker
from sagemaker.huggingface import HuggingFace

# gets role for executing training job
role = sagemaker.get_execution_role()
hyperparameters = {
    'model_name_or_path':'Helsinki-NLP/opus-mt-de-en',
    'output_dir':'/opt/ml/model',
    'dataset_name':  'wmt14', 
    'source_lang': 'de', 
    'target_lang': 'en',
    'dataset_config_name': 'de-en'
}


# git configuration to download our fine-tuning script
git_config = {
    'repo': 'https://github.com/huggingface/transformers.git',
    'branch': 'v4.36.0'}


# creates Hugging Face estimator
huggingface_estimator = HuggingFace(
	entry_point='run_translation.py',
	source_dir='./examples/pytorch/translation',
	instance_type='ml.g4dn.xlarge',
	instance_count=1,
	role=role,
	git_config=git_config,
	transformers_version='4.36.0',
	pytorch_version='2.1.0',
	py_version='py310',
	hyperparameters = hyperparameters,
)

# starting the train job
huggingface_estimator.fit()
\end{lstlisting}
\caption{Working Sample Code to Fine-tune a Translation model with Huggingface Transformers}
\label{working-sample-fig}
\end{figure}

\clearpage
\twocolumn

\newpage
\onecolumn
\begin{figure}[!ht]
\begin{lstlisting}[language=Python]
import sagemaker
from sagemaker.huggingface import HuggingFace
from sagemaker.tuner import IntegerParameter
from sagemaker.tuner import HyperparameterTuner


# gets role for executing training job
role = sagemaker.get_execution_role()
hyperparameters = {
  'model_name_or_path':'Helsinki-NLP/opus-mt-de-en',
  'output_dir':'/opt/ml/model',
  'dataset_name':  'wmt14', 
  'source_lang': 'de', 
  'target_lang': 'en',
  'dataset_config_name': 'de-en',
}


metric_definitions=[
  {"Name":  "train_loss", "Regex":  "'train_loss': ([0-9]+(.|e\-)[0-9]+),?"}
]


# creates Hugging Face estimator
huggingface_estimator = HuggingFace(
  entry_point='my_run_translations.py',
  source_dir='./scripts',
  instance_type='ml.g4dn.xlarge',
  instance_count=1,
  role=role,
  transformers_version='4.36.0',
  pytorch_version='2.1.0',
  py_version='py310',
  metric_definitions=metric_definitions,
  hyperparameters = hyperparameters,
)

hyperparameter_ranges = {
    "max_source_length": IntegerParameter(80,200),
    "max_target_length": IntegerParameter(80,200),
}
objective_metric_name = "max_source_length"
objective_type = "Maximize"

hpo_definitons = [
  {"Name":  "max_source_length", 
   "Regex": '\"max_source_length\"\:\s([-+]?[0-9]*\.?[0-9]+(?:[eE][-+]?[0-9]+)?).*'}
]

tuner = HyperparameterTuner(
  huggingface_estimator,
  objective_metric_name,
  hyperparameter_ranges,
  hpo_definitons,
  max_jobs=10,
  max_parallel_jobs=1,
  objective_type=objective_type,
)

tuner.fit()
\end{lstlisting}
\caption{Working Sample Code to do Hyperparameter Tuning for a Translation model with Huggingface Transformers}
\label{fine-tuning-sample}

\end{figure}
\clearpage
\twocolumn

\end{document}